\documentclass[wcp]{jmlr}
\usepackage{newtxtext}

\usepackage[table]{xcolor}
\usepackage{pifont}
\usepackage{enumitem}
\usepackage{array}
\usepackage{pgfplots}
\usepackage{comment}
\usepackage{multirow}
\usepackage{array}
\usepackage{rotating}

\usepackage{booktabs}
\usepackage{array}
\usepackage{threeparttable}
\usepackage{fontawesome5}

\usepackage{longtable}

\usepackage{booktabs}

\usepackage{lineno}

\pgfplotsset{compat=1.18}
\usepgfplotslibrary{groupplots}
\usepgfplotslibrary{colormaps}
\pgfplotsset{
  colormap={Blues}{rgb255=(247,251,255) rgb255=(198,219,239) rgb255=(107,174,214) rgb255=(33,113,181) rgb255=(8,48,107)}
}

\usepackage[most]{tcolorbox}
\usepackage{xcolor}
\usepackage{listings}

\definecolor{promptbg}{RGB}{248,249,251}
\definecolor{promptframe}{RGB}{60,89,130}
\definecolor{prompttitle}{RGB}{31,54,87}

\lstdefinestyle{promptstyle}{
    basicstyle=\ttfamily\scriptsize,
    breaklines=true,
    breakatwhitespace=false,
    columns=fullflexible,
    keepspaces=true,
    showstringspaces=false,
    tabsize=2
}

\newtcblisting{promptbox}{
    enhanced,
    breakable,
    colback=promptbg,
    colframe=promptframe,
    coltitle=white,
    title=\textbf{Representative Prompt for Aspect-Oriented Evidence Extraction},
    fonttitle=\bfseries,
    listing only,
    listing style=promptstyle,
    arc=1.2mm,
    boxrule=0.7pt,
    left=1.5mm,
    right=1.5mm,
    top=1.2mm,
    bottom=1.2mm,
    before skip=8pt,
    after skip=8pt
}

\makeatletter
\let\Ginclude@graphics\@org@Ginclude@graphics 
\makeatother

\jmlryear{2026}
\jmlrworkshop{ACML 2026}

\usepackage{hyperref}

\title{\textsc{MAPLE}: Medical Aspect-Based Summarization with Phrase-Level Evidence}

\author{
\textbf{Bohao Chu}\textsuperscript{\normalfont 1,*, $^\dagger$} \;
\textbf{Hendrik Damm}\textsuperscript{\normalfont 2,5,$^\dagger$} \;
\textbf{Tabea M. G. Pakull}\textsuperscript{\normalfont 4,2} \;
\textbf{Sameh Frihat}\textsuperscript{\normalfont 1} \;
\textbf{Georg Lodde}\textsuperscript{\normalfont 3} \;  \\
\textbf{Elisabeth Livingstone}\textsuperscript{\normalfont 3} \;
\textbf{Christoph M. Friedrich}\textsuperscript{\normalfont 2,5} \;
\textbf{Norbert Fuhr}\textsuperscript{\normalfont 1} \\
\addr\textsuperscript{1}University of Duisburg-Essen,
\addr\textsuperscript{2}University of Applied Sciences and Arts Dortmund,  \\
\addr\textsuperscript{3}University Hospital Essen,
\addr\textsuperscript{4}Institute for Transfusion Medicine, University Hospital Essen, \\
\addr\textsuperscript{5}Institute for Medical Informatics, Biometry and Epidemiology (IMIBE) \\
{
\textsuperscript{*} Correspondence: \href{mailto:bohao.chu@qq.com}{bohao.chu@qq.com}
} \; $^\dagger$ Equal contribution
}

\begin{document}

\makeatletter
\let \@jmlrpages \@empty
\makeatother

\maketitle
\pagestyle{empty}
\thispagestyle{empty}

\begin{abstract}
Trustworthy clinical summarization requires every claim to be traceable to its evidence, yet existing attribution often resolves only to the sentence or document, leaving clinicians to scan surrounding text for the few words that matter. We argue that the unit of attribution should match the unit of verification: the precise phrase the reader's eye must land on. We present \textsc{MAPLE} (\textbf{M}edical \textbf{A}spect-Based Summarization with \textbf{P}hrase-\textbf{L}evel \textbf{E}vidence), a human-annotated benchmark that grounds each summarized claim in both cited sentences and contributory phrases within them. Spanning $152$ randomized controlled trial (RCT) abstracts and $16$ clinically motivated aspects, \textsc{MAPLE} comprises $1{,}799$ aspect-based summaries with two-level evidence. We further introduce a decoupled evaluation framework that separately scores content, traceability, and locatability, together with a proxy for the amount of source text a clinician must inspect to verify a claim. Benchmarking eleven LLMs shows that sentence-level citation is consistently strong (C-F1 up to $90.9\%$), while phrase-level grounding remains less stable and the most discriminative axis across models (P-F1 $66.1$--$84.5\%$). These results suggest that the key challenge is not only producing accurate summaries, but localizing their supporting evidence precisely enough for efficient clinical verification. Data and code are available at \url{https://github.com/chubohao/maple}.
\end{abstract}

\section{Introduction}
Aspect-based summarization condenses a document into several aspect-specific summaries, each targeting a single facet of interest~\citep{yang-etal-2023-oasum,takeshita-etal-2024-aclsum}. The need is acute in medicine, where a clinician reading a randomized controlled trial (RCT) must extract structured evidence on participants, interventions, outcomes, and adverse events, often across many trials at once, to synthesize findings and inform decisions at the point of care~\citep{chu-etal-2025-tracsum}.

Large language models (LLMs) make such summaries easy to produce, yet their fluency conceals a verification problem: a generated summary can hallucinate or drift subtly from the facts~\citep{mallen-etal-2023-trust}, a failure that becomes consequential when a misreported dosage or adverse-event grade may alter treatment. Context attribution mitigates this by linking each claim back to its source, letting the reader check it against the evidence~\citep{xie-etal-2024-doclens}. Existing methods, however, attribute only at the document or sentence level, still forcing the clinician to scan a paragraph for the one phrase that supports a given claim~\citep{gao-etal-2023-enabling,slobodkin-etal-2024-attribute}.

\vspace{0.1cm}
\noindent\textbf{Why phrase level.}
Attribution in a high-stakes setting must be both \emph{traceable}, so that every claim can be audited against a specific source, and \emph{locatable}, so that the audit is fast. Sentence-level citation delivers the first but not the second: the clinician knows where to look, yet still reads through a paragraph to find the words that matter. The right unit of attribution is thus the unit of verification, the contiguous phrase the eye must land on, such as a dose, a hazard ratio, or an event rate. We close this granularity gap by studying \textit{phrase-level} context attribution, aligning each summarized claim with the phrases within its cited sentences on which it rests and can be verified at a glance.

\vspace{0.1cm}
\noindent\textbf{Contributions.} Our contribution is threefold. \textbf{(1)} A phrase-level evidence benchmark: $152$ RCT abstracts across $16$ clinically motivated aspects, yielding $1{,}799$ applicable aspect-based summaries, each paired with cited sentences and contributory phrases (\S\ref{sec:dataset}). \textbf{(2)} A decoupled evaluation that independently scores content, traceability, and locatability across seven metrics, complemented by a verification-cost proxy that localizes failures to a single axis (\S\ref{sec:benchmark}). \textbf{(3)} An empirical study showing that traceability is comparatively strong while locatability remains the dominant bottleneck, and that scale improves grounding faster than content within a fixed evaluation protocol (\S\ref{sec:experiment}).

\begin{table}[t]
\centering
\fontsize{9pt}{8.3pt}\selectfont
\setlength{\tabcolsep}{8pt}
\renewcommand{\arraystretch}{1.15}
\caption{Comparison of \textsc{MAPLE} with existing aspect-based and evidence-grounded summarization resources. Attribution denotes explicit sentence- or phrase-level evidence. $^\dagger$ACLSum provides sentence highlights but does not frame them as citations.}
\vspace{-5pt}
\begin{tabular}{llllll}
\toprule
\textbf{Benchmark} & \textbf{Domain} & \textbf{Aspects} & \textbf{Inst.} & \textbf{Annotation} & \textbf{Attribution} \\
\midrule
OASUM~\citep{yang-etal-2023-oasum}        & open          & open          & 3.7M     & automatic & -- \\
ACLSum~\citep{takeshita-etal-2024-aclsum} & sci.\ papers  & 3             & 250      & human     & sentence$^\dagger$ \\
FactPICO~\citep{joseph-etal-2024-factpico}& medical       & 4  & 345      & LLM-gen.  & -- \\
TracSum~\citep{chu-etal-2025-tracsum}     & medical       & 7             & 3{,}500  & LLM+Human & sentence \\
\midrule
\textbf{Ours}                            & medical (RCT) & \textbf{16}   & \textbf{1{,}799} & human & \textbf{sentence + phrase} \\
\bottomrule
\end{tabular}
\label{tab:related}
\vspace{-15pt}
\end{table}

\section{Related Work}
\noindent\textbf{Aspect-Based and Medical Summarization.}
Aspect-based summarization generates one summary per user-specified aspect~\citep{hayashi-etal-2021-wikiasp,zhang-etal-2023-macsum}, later extended to open-aspect settings~\citep{yang-etal-2023-oasum} and to scholarly documents such as ACLSum~\citep{takeshita-etal-2024-aclsum}. These target general or scientific prose, saying little about settings where summaries must be at once aspect-structured and verifiable. In medicine, structured summarization has centered on the PICO framework~\citep{richardson1995well}, from sequence-labeling extractors~\citep{jin-szolovits-2018-pico} to LLM-era audits of plain-language summaries~\citep{joseph-etal-2024-factpico}. Across these lines, aspects remain coarse, often limited to PICO elements, and evidence links mostly implicit. We extend both dimensions, covering $16$ clinically motivated aspects and grounding every summary in cited sentences and contributory phrases so that every claim remains independently auditable.

\vspace{0.1cm}
\noindent\textbf{Context Attribution.}
Recent work links generated content back to its source so readers can verify each claim, by emitting citations during generation~\citep{menick2022teaching,gao-etal-2023-enabling}, aligning outputs to source spans post hoc~\citep{li2016understanding,honovich-etal-2022-true-evaluating}, or conditioning on retrieved evidence~\citep{slobodkin-etal-2024-attribute,chu-etal-2025-tracsum}. These provide \emph{traceability} but not \emph{locatability}: a document- or sentence-level citation still forces the reader to scan substantial text. We push attribution to the phrase level. Table~\ref{tab:related} situates \textsc{MAPLE} among the closest resources.

\section{Task Formulation} \label{sec:task}
We formulate medical aspect-based summarization with phrase-level context attribution as a structured generation task. Given an article and a target aspect, a system must produce an aspect-specific summary with two levels of grounding: cited sentences that make it \emph{traceable} to the source, and contributory phrases within those sentences that make it \emph{locatable} at a glance. This phrase layer distinguishes our task from both standard aspect-based summarization and citation-augmented generation, tying each claim to a concrete span rather than the surrounding paragraph.

\vspace{0.1cm}
\noindent\textbf{Inputs and outputs.}
Let $d = [c_1, \ldots, c_{|d|}]$ be a source abstract given as a sequence of uniquely indexed sentences, and $a \in \mathcal{A}$ a target aspect drawn from a predefined schema of $|\mathcal{A}| = 16$ clinically motivated aspects (Table~\ref{tab:aspects}). A system $\mathcal{M}$ maps
\begin{equation}
\mathcal{M}: (d, a) \;\mapsto\; (\textit{sum}',\; \mathcal{C}',\; \mathcal{P}'),
\end{equation}
where $\textit{sum}'$ summarizes aspect $a$ in $d$, $\mathcal{C}' \subseteq \{c_1, \ldots, c_{|d|}\}$ is the set of cited sentences, and $\mathcal{P}'$ the set of contributory phrases, each a contiguous span of some $c \in \mathcal{C}'$ reflected, verbatim or by paraphrase, in $\textit{sum}'$ for rapid verification by a time-pressed clinician.

\vspace{0.1cm}
\noindent\textbf{Grounding constraints.}
A valid output must satisfy: (i) every cited sentence belongs to $d$; (ii) every contributory phrase is a contiguous substring of a cited sentence, not free-form text; and (iii) the summary is supported by $\mathcal{C}'$ and faithfully reflects $\mathcal{P}'$. Constraint (ii) is what separates phrase-level grounding from free-text rationales: a phrase must be a verbatim span of the source, so it can be highlighted in place and audited without trusting a model-generated paraphrase. If $d$ holds no information relevant to $a$, the system abstains, returning $\textit{sum}' = $ empty and $\mathcal{C}' = \mathcal{P}' = \emptyset$, which lets the framework reward correct silence rather than penalizing it as a costly omission.

\section{The Dataset} \label{sec:dataset}
\subsection{Source Corpora and Aspect Schema} \label{sec:source}
We restrict the corpus to melanoma RCTs. This focus is deliberate: collaboration with dermatologists at University Hospital Essen guides the annotation, while a single therapeutic area keeps abstracts structurally comparable and clinical guidance consistent. RCTs also sit among the highest tiers of evidence~\citep{higgins2024cochrane}. From PubMed,\footnote{\url{https://pubmed.ncbi.nlm.nih.gov} (last accessed May~1,~2026).} we collect abstracts that (i) concern melanoma, (ii) appeared within the past decade, (iii) are in English, (iv) are classified as RCTs, and (v) appeared in Q1/Q2 journals according to JCR category quartiles~\citep{clarivate2025jcr}, and randomly sample $152$ of them for annotation under a fixed budget. We use abstracts rather than full texts because they are consistently retrievable through PubMed and compactly report key trial aspects, a practice common in medical summarization~\citep{joseph-etal-2024-factpico,chu-etal-2025-tracsum}. Drawing on clinician interviews and the PICO framework~\citep{richardson1995well}, we define an aspect schema $\mathcal{A}$ of $16$ facets commonly reported in RCT abstracts, grouped into Population \& Design, Intervention, Objective \& Outcomes, and Safety \& Meta-information. For each aspect, we specify a \textit{minimal reporting requirement}: the essential fields a summary should contain when the aspect applies, which standardizes annotation and makes summaries comparable across trials. Table~\ref{tab:aspects} lists the schema and worked examples.

\begin{table*}[t!]
\centering
\fontsize{9pt}{12pt}\selectfont
\setlength{\tabcolsep}{9pt}
\renewcommand{\arraystretch}{1}
\caption{The sixteen medical aspects predefined in \textsc{MAPLE}, grouped by their role in RCT reporting. Each example illustrates a one-sentence summary, with the parts that reflect contributory phrases shown in \textbf{bold} and the minimal reporting requirement given in [brackets].}
\vspace{-5pt}
\begin{tabular}{@{}l@{\hspace{36pt}}l l@{}}
\toprule
\textbf{Code} & \textbf{Aspect} & \textbf{Example summary} \\
\midrule
\multicolumn{3}{@{}l}{\textit{Population \& Design}} \\
\arrayrulecolor{gray}\hline\arrayrulecolor{black}
P  & Participants       & \textbf{563} [number] patients with \textbf{metastatic melanoma} [disease]. \\
R  & Randomization      & \textbf{Randomized 2:1} [ratio] to \textbf{binimetinib or dacarbazine} [group]. \\
B  & Blinding           & The study was \textbf{double-blind} [blinding]. \\
\arrayrulecolor{gray}\hline\arrayrulecolor{black}
\multicolumn{3}{@{}l}{\textit{Intervention}} \\
\arrayrulecolor{gray}\hline\arrayrulecolor{black}
I  & Intervention       & \textbf{Dabrafenib (150 mg) plus trametinib (2 mg)} [administration]. \\
C  & Comparator         & \textbf{Ipilimumab monotherapy} [name] at \textbf{3 mg/kg} [administration]. \\
M  & Medicines          & \textbf{Dabrafenib} and \textbf{trametinib} [medicine]. \\
TD & Treatment Duration & Administered \textbf{until disease progression} [condition]. \\
\arrayrulecolor{gray}\hline\arrayrulecolor{black}
\multicolumn{3}{@{}l}{\textit{Objective \& Outcomes}} \\
\arrayrulecolor{gray}\hline\arrayrulecolor{black}
OB & Objective          & To determine the \textbf{5-year survival rates} [item] under \textbf{BRAF/MEK} [treatment]. \\
O  & Outcomes           & \textbf{5-year OS rate} [endpoint] of \textbf{34\%} [value]. \\
F  & Findings           & \textbf{Improved clinical outcomes} [finding] over monotherapy. \\
PE & Primary Endpoint   & \textbf{Progression-free survival} [endpoint]. \\
SE & Secondary Endpoint & \textbf{Long-term survival rates} [endpoint]. \\
FD & Follow-Up Duration & Median follow-up of \textbf{22 months} [duration]. \\
\arrayrulecolor{gray}\hline\arrayrulecolor{black}
\multicolumn{3}{@{}l}{\textit{Safety \& Meta-information}} \\
\arrayrulecolor{gray}\hline\arrayrulecolor{black}
AE & Adverse Events     & Most common was \textbf{headache} [event] \textbf{(55.3\%)} [value]. \\
FU & Funding            & Funded by \textbf{Bristol-Myers Squibb} [sponsor]. \\
RE & Registration       & ClinicalTrials.gov number \textbf{NCT03698019} [number]. \\
\bottomrule
\end{tabular}
\vspace{-15pt}
\label{tab:aspects}
\end{table*}

\subsection{Annotation Protocol} \label{sec:protocol}
A medical student from a professional biomedical Annotation Lab Essen annotated all $152$ articles through a custom web interface, at the local student-assistant rate. Guidelines were developed and reviewed by dermatologists at University Hospital Essen, who also resolved clinical ambiguities flagged by the annotator. Annotation proceeded in three phases. \textbf{Phase~I} assigns each sentence zero or more aspect labels from $\mathcal{A}$. \textbf{Phase~II} writes, for each applicable aspect, a summary from its Phase-I sentences satisfying the minimal reporting requirement. \textbf{Phase~III} marks the contributory phrases within those sentences that the summary reflects, verbatim or by paraphrase. To support \S\ref{sec:quality}, a second annotator from the same lab independently re-labeled a stratified random sample of $20$ articles (seed~$=42$, $239$ aspect instances, ${\sim}13\%$ of the corpus) under the same protocol.

\subsection{Quality Assessment} \label{sec:quality}
Following common practice in benchmark construction under finite annotation budgets~\citep{wang-etal-2020-asking, bhandari-etal-2020-evaluating}, we assess quality from two complementary angles: annotator \emph{agreement} on a double-annotated subset, and \emph{expert review} of the broader corpus.

\vspace{0.1cm}
\noindent\textbf{Inter-annotator agreement.} On the $20$-article subset, we score all three phases: Phase~I (sentence selection) with Cohen's $\kappa$ and F1 over per-sentence binary labels; Phase~II (summary writing) with ROUGE-L F1~\citep{lin-2004-rouge} and BERTScore F1~\citep{zhang2020bertscore} (RoBERTa-large, baseline-rescaled); Phase~III (phrase selection) with token-level F1 within the parent sentence. Mutual abstention counts as perfect agreement. Averaged over the $16$ aspects (Table~\ref{tab:iaa}), sentence selection reaches $\kappa=0.80$, in the ``substantial'' range~\citep{landis1977}; summaries reach ROUGE-L $=0.67$ and BERTScore $=0.72$; phrases reach token-F1 $=0.64$, where the residual gap reflects span-boundary subjectivity rather than disagreement on what is contributory, motivating the semantic matching of \S\ref{sec:eval-phrase}. Difficulty tracks aspect structure: near-verbatim aspects (Blinding, Funding, Registration) reach $\kappa\geq 0.93$, while Comparator is weakest ($\kappa=0.44$), reflecting genuine ambiguity in combination-therapy trials where intervention and comparator arms are not always clearly demarcated, so its low agreement signals task difficulty rather than annotator error.

\vspace{0.1cm}
\noindent\textbf{Independent human review.} Agreement quantifies consistency but cannot fully assess whether annotations are faithful and minimally redundant, so we complement it with independent human review. Two reviewers, blind to the annotators and to each other, rated a stratified sample of $50$ articles (seed~$=42$) on a 5-point Likert scale along two dimensions, \textit{completeness} (whether the annotation captures the target aspect) and \textit{conciseness} (whether it avoids irrelevant content), at each of the three layers. Scores exceed $4.6$ on both dimensions for every layer, averaged over the $16$ aspects, with high reviewer consistency: $92.1\%$ exact match, $97.4\%$ within one point, and mean absolute error $0.109$. The two streams converge: the corpus is consistently labeled and clinically faithful, even where double annotation was not feasible at full corpus scale.

\begin{table*}[t]
\centering
\fontsize{9.5pt}{9.5pt}\selectfont
\setlength{\tabcolsep}{9.5pt}
\renewcommand{\arraystretch}{1}
\caption{Inter-annotator agreement on the double-annotated subset, macro-averaged by aspect group. The three phases correspond to sentence (I), summary (II), and phrase (III) annotation.}
\vspace{-5pt}
\label{tab:iaa}
\begin{tabular}{llccccc}
\toprule
& \textbf{Metric} & \textbf{Pop.\,\&\,Des.} & \textbf{Interv.} & \textbf{Obj.\,\&\,Out.} & \textbf{Safety\,\&\,Meta} & \textbf{Average} \\
\midrule
\multirow{2}{*}{Phase~I}  & $\kappa$    & 0.869 & 0.681 & 0.768 & 0.965 & \textbf{0.802} \\
                          & F1          & 0.912 & 0.767 & 0.806 & 0.970 & \textbf{0.847} \\
\midrule
\multirow{2}{*}{Phase~II} & ROUGE-L     & 0.677 & 0.638 & 0.641 & 0.770 & \textbf{0.671} \\
                          & BERTScore   & 0.700 & 0.667 & 0.691 & 0.840 & \textbf{0.715} \\
\midrule
Phase~III                 & Token-F1    & 0.637 & 0.583 & 0.576 & 0.851 & \textbf{0.641} \\
\bottomrule
\end{tabular}
\vspace{-15pt}

\end{table*}

\subsection{Corpus Statistics} \label{sec:stats}
Figure~\ref{fig:corpus_stats} summarizes the corpus statistics. The $152$ abstracts average $306.5$ words and $13.6$ sentences, with lengths skewed toward shorter texts (Figure~\ref{fig:corpus_stats}a,b). In contrast, the $1{,}799$ aspect instances are much more compact: each summary averages $20.9$ words (Figure~\ref{fig:corpus_stats}c), $1.44$ cited sentences (Figure~\ref{fig:corpus_stats}d), and $3.4$ contributory phrases (Figure~\ref{fig:corpus_stats}e). This gap between a ${\sim}14$-sentence abstract and a small evidence set highlights the task: systems must identify precise supporting evidence from a much larger source, exactly the challenge targeted by phrase-level attribution.

\vspace{0.1cm}
\noindent\textbf{Aspect coverage and placement.}
Coverage is uneven (Figure~\ref{fig:corpus_stats}f). Core aspects are near-universal, such as \textit{Participants} ($n{=}152$), \textit{Intervention} ($n{=}150$), \textit{Outcomes} ($n{=}150$), and \textit{Findings} ($n{=}151$), while several appear in fewer than half the articles, notably \textit{Secondary Endpoints} ($n{=}57$), \textit{Blinding} ($n{=}61$), and \textit{Funding} ($n{=}42$). This skew is informative: aspects bearing on trial interpretation and reproducibility are not uniformly reported, yielding a realistic mix of frequent, sparse, and absent evidence cases. We also observe positional regularities, with \textit{Objective} evidence near the start of abstracts and \textit{Findings} near the end while others are more dispersed; model difficulty in turn varies substantially across aspect types, a pattern our per-aspect analysis confirms (\S\ref{sec:exp-aspect}).

\begin{figure}[t]
\centering
\pgfplotsset{
  compat=1.18,
  histstyle/.style={
    ybar, ymin=0, bar width=6.5pt,
    width=0.53\textwidth, height=3.35cm,
    tick align=outside, tickpos=left,
    major tick length=2pt, tick style={line width=0.4pt},
    legend image code/.code={\draw[fill=blue,draw=blue] (0cm,-0.08cm) rectangle (0.18cm,0.12cm);},
    legend style={at={(0.5,1.03)}, anchor=south, draw=none, font=\scriptsize, fill=none, inner sep=1pt},
    label style={font=\scriptsize}, tick label style={font=\scriptsize},
    ymajorgrids, grid style={gray!15},
  }
}
\begin{tikzpicture}
\begin{groupplot}[
    group style={group size=2 by 3, horizontal sep=1.3cm, vertical sep=1.1cm},
    histstyle,
]

\nextgroupplot[ybar, bar width=8pt, ymax=65, xmin=100, xmax=680, xtick={200,300,400,500,600}]
\addplot[fill=blue, draw=blue] coordinates {(125,2)(175,8)(225,17)(275,59)(325,38)(375,10)(425,2)(475,5)(525,8)(575,1)(625,2)};
\addlegendentry{Word Count In Abstract (a)}
\node[font=\scriptsize] at (axis description cs:0.5,-0.34) {(a)};

\nextgroupplot[
    ybar,
    bar width=5pt,
    ymax=35,
    xmin=3.5, xmax=24.5,
    xtick={4,6,8,10,12,14,16,18,20,22,24},
    xticklabel style={font=\tiny}
]
\addplot[fill=blue, draw=blue] coordinates {
    (4,1)
    (6,2)
    (7,1)
    (8,4)
    (9,3)
    (10,7)
    (11,16)
    (12,29)
    (13,25)
    (14,20)
    (15,12)
    (16,3)
    (17,11)
    (18,3)
    (19,3)
    (20,3)
    (21,3)
    (22,4)
    (23,1)
    (24,1)
};
\addlegendentry{Sentence Count In Abstract (b)}
\node[font=\scriptsize] at (axis description cs:0.5,-0.34) {(b)};

\nextgroupplot[
    ybar,
    bar width=7pt,
    ymax=600,
    xmin=0, xmax=110,
    xtick={0,20,40,60,80,100},
    ytick={0,200,400,600}
]
\addplot[fill=blue, draw=blue] coordinates {
    (5,455)
    (15,483)
    (25,438)
    (35,259)
    (45,93)
    (55,38)
    (65,22)
    (75,5)
    (85,4)
    (95,1)
    (105,1)
};
\addlegendentry{Word Count In Summary (c)}
\node[font=\scriptsize] at (axis description cs:0.5,-0.34) {(c)};

\nextgroupplot[
    ybar,
    bar width=11pt,
    ymax=1500,
    xmin=0.4, xmax=6.6,
    xtick={1,2,3,4,5,6},
    ytick={0,500,1000,1500}
]
\addplot[fill=blue, draw=blue] coordinates {
    (1,1221)
    (2,418)
    (3,121)
    (4,26)
    (5,11)
    (6,2)
};
\addlegendentry{Count of Sentence Cited By Summary (d)}
\node[font=\scriptsize] at (axis description cs:0.5,-0.34) {(d)};

\nextgroupplot[ybar, bar width=4pt, ymax=600, xmin=0, xmax=24, xtick={1,5,10,15,20}, ytick={0,200,400,600}]
\addplot[fill=blue, draw=blue] coordinates {(1,525)(2,412)(3,235)(4,201)(5,118)(6,93)(7,48)(8,48)(9,30)(10,24)(11,14)(12,13)(13,5)(14,6)(15,7)(16,7)(17,7)(18,1)(19,1)(20,2)(21,1)(23,1)};
\addlegendentry{Count of Phrase Cited By Summary (e)}
\node[font=\scriptsize] at (axis description cs:0.5,-0.34) {(e)};
\nextgroupplot[ybar, bar width=6pt, ymin=40, ymax=250,
    xmin=0.4, xmax=16.6,
    xtick={1,2,3,4,5,6,7,8,9,10,11,12,13,14,15,16},
    xticklabels={OB,P,I,O,M,C,R,TD,PE,SE,AE,F,FD,B,FU,RE},
    x tick label style={font=\tiny}, ytick={40,80,120,160},
    nodes near coords, nodes near coords style={rotate=90, anchor=west, font=\tiny},
    every node near coord/.append style={/pgf/number format/precision=0, /pgf/number format/fixed}]
\addplot[fill=blue, draw=blue] coordinates {(1,138)(2,152)(3,150)(4,150)(5,140)(6,134)(7,126)(8,95)(9,121)(10,57)(11,111)(12,151)(13,85)(14,61)(15,42)(16,86)};
\addlegendentry{Abstract Coverage (f)}
\node[font=\scriptsize] at (axis description cs:0.5,-0.34) {(f)};
\end{groupplot}
\end{tikzpicture}
\vspace{-25pt}
\caption{Corpus statistics for \textsc{MAPLE}: abstract and summary lengths, citation and phrase counts, and per-aspect coverage across 1,799 instances. Aspect codes follow \autoref{tab:aspects}.}
\label{fig:corpus_stats}
\vspace{-20pt}
\end{figure}

\section{Evaluation Framework} \label{sec:benchmark}
A system outputs a three-tuple $(\textit{sum}', \mathcal{C}', \mathcal{P}')$ whose components fail independently: the summary may be fluent yet incomplete, the citations misplaced, or the phrases too broad to localize the evidence. We therefore evaluate three axes: \emph{content} (whether the summary is correct), \emph{traceability} (whether its claims are backed by cited sentences), and \emph{locatability} (whether the cited evidence is narrowed to the relevant phrases). Recall and precision use direction-specific predicates throughout.

\subsection{Preliminaries} \label{sec:eval-prelim}
The framework relies on three swappable components: a decomposition model $\mathcal{E}$ that maps a summary $\textit{sum}$ to atomic subclaims $\mathcal{L}_{\textit{sum}} = \{l_1, \ldots, l_k\}$; an entailment model $\phi$, where $p \models h$ if and only if $\phi(p,h)=1$; and a sentence encoder with cosine similarity $\mathcal{S}(\cdot,\cdot)$ for paraphrase-tolerant matching. We instantiate them with greedy-decoded \textit{Qwen3.6-35B-A3B}~\citep{qwen36_35b_a3b}, TRUE~\citep{honovich-etal-2022-true-evaluating}, and \textit{all-MiniLM-L6-v2}~\citep{wang2020minilm}, respectively. The decomposer is fixed and external to the evaluated systems; the judges are validated against human labels in \S\ref{sec:exp-metric}, where component swaps leave the rankings stable.

\vspace{0.1cm}
\noindent\textbf{Abstention.}
If an aspect is absent, the reference contains no summary, citations, or phrases. Recall is set to $1$ for empty references, while empty hypotheses receive credit only when the corresponding reference side is also empty; otherwise they count as omissions. This rewards correct silence while penalizing missed or fabricated evidence under uneven aspect coverage (\S\ref{sec:stats}).

\subsection{Content Axis} \label{sec:eval-summary}
Following \textsc{DocLens}~\citep{xie-etal-2024-doclens} and \textsc{TracSum}~\citep{chu-etal-2025-tracsum}, we score content over atomic subclaims, making the metric robust to surface paraphrase. Let $\mathcal{L}_{\textit{sum}}$ and $\mathcal{L}'_{\textit{sum}}$ be the reference and generated subclaim sets; a subclaim is valid if entailed by the opposing summary. Recall and precision are $\textbf{S-R} = |\{l \in \mathcal{L}_{\textit{sum}} : \textit{sum}' \models l\}| / |\mathcal{L}_{\textit{sum}}|$ and $\textbf{S-P} = |\{l' \in \mathcal{L}'_{\textit{sum}} : \textit{sum} \models l'\}| / |\mathcal{L}'_{\textit{sum}}|$, measuring whether reference subclaims are recovered and whether generated subclaims are supported, respectively: omissions lower \textbf{S-R}, hallucinations lower \textbf{S-P}, and neither can inflate the score. Since both directions share the decomposer, any decomposition bias is shared across systems. This axis measures whether a subclaim is \emph{true}; its \emph{traceability} is evaluated next.

\subsection{Traceability Axis} \label{sec:eval-citation}
The traceability axis evaluates evidence from two directions: whether each cited sentence is warranted, and whether each generated claim is backed, both through the single-premise primitive $c \models l'$. For recall, a reference citation is covered if and only if the system cites the same sentence \citep{chu-etal-2025-tracsum}: $\textbf{C-R} = |\mathcal{C} \cap \mathcal{C}'| / |\mathcal{C}|$. For precision, we adopt a reference-free criterion: a system citation is credited if it supports at least one generated subclaim, so a sentence the expert did not cite is still rewarded when it genuinely supports the summary: $\textbf{C-P} = |\{c \in \mathcal{C}' : \exists\, l' \in \mathcal{L}'_{\textit{sum}},\, c \models l'\}| / |\mathcal{C}'|$. These two directions miss a complementary failure: a subclaim may be true yet unsupported by any cited sentence. We therefore add an attribution score from claims back to citations, where $l'$ is attributable if some cited sentence entails it: $\textbf{C-A} = |\{l' \in \mathcal{L}'_{\textit{sum}} : \exists\, c \in \mathcal{C}',\, c \models l'\}| / |\mathcal{L}'_{\textit{sum}}|$. Together the three metrics separate traceability failures: irrelevant evidence lowers \textbf{C-P}, missed expert evidence lowers \textbf{C-R}, and untraceable claims lower \textbf{C-A}. We keep \textbf{C-A} separate from \textbf{S-P} because a claim can be true without being traceable to the system's own evidence, a gap that no other metric captures and one that directly undermines a clinician's ability to audit the claim.

\subsection{Locatability Axis} \label{sec:eval-phrase}

\noindent\textbf{Two admissibility gates.}
Locatability evaluates whether a system narrows each cited sentence to the spans a clinician must inspect. Since lexical overlap misses paraphrases (e.g., ``administered intravenously'' vs.\ ``given by IV''), we match phrases by encoder cosine similarity $\mathcal{S}(\cdot,\cdot)$ with threshold $\tau$. A predicted phrase $p'$ must pass two gates. \textbf{Provenance} requires $p'$ to be a contiguous span of a cited sentence:
\begin{equation}
\fontsize{9pt}{9pt}\selectfont
\operatorname{prov}(p') =
\mathbb{I}\!\left[\,p' \sqsubseteq c \text{ for some } c \in \mathcal{C}'\,\right],
\end{equation}
where $\sqsubseteq$ denotes substring containment after lightweight text normalization. \textbf{Faithfulness} requires $p'$ to be reflected in the generated summary, either verbatim or by paraphrase. To prevent over-broad spans from passing through sentence-level similarity alone, we require most tokens in $p'$ to match short summary windows:
\begin{equation}
\fontsize{8pt}{9pt}\selectfont
\operatorname{faith}(p') =
\mathbb{I}\!\left[
\frac{1}{|\mathcal{T}(p')|}
\sum_{t \in \mathcal{T}(p')}
\mathbb{I}\!\left[
\max_{w \in \mathcal{W}(\textit{sum}',t)}
\mathcal{S}(t,w) \geq \tau
\right]
\geq \eta
\right],
\end{equation}
where $\mathcal{T}(p')$ is the token sequence of $p'$, and $\mathcal{W}(\textit{sum}',t)$ denotes contiguous windows of one to three tokens in $\textit{sum}'$. Thus, each phrase token is matched against short summary windows rather than isolated words. We set $\tau=0.7$ and $\eta=0.8$ as fixed, untuned defaults and evaluate on all $1{,}799$ instances; their effect is analyzed in the threshold sweep of \S\ref{sec:exp-metric}, which also reports a hard lexical-containment variant isolating the effect of paraphrase tolerance.

\vspace{0.1cm}
\noindent\textbf{Precision and recall.} Letting $\mathcal{P}$ and $\mathcal{P}'$ be the reference and predicted phrase sets, precision is $\textbf{P-P} = |\{p' \in \mathcal{P}' : \operatorname{prov}(p') \wedge \operatorname{faith}(p')\}| / |\mathcal{P}'|$ and recall is $\textbf{P-R} = |\{p \in \mathcal{P} : \exists\, p' \in \mathcal{P}',\, \operatorname{prov}(p') \wedge \mathcal{S}(p', p) \geq \tau\}| / |\mathcal{P}|$. \textbf{P-P} measures whether predicted phrases are admissible and faithful, while \textbf{P-R} measures coverage of the expert set. A wider span may raise \textbf{P-R} by covering several reference phrases, but it risks lowering \textbf{P-P} through the faithfulness gate and is penalized by the cost proxy below, so highlighting whole sentences inflates coverage rather than tightening localization.

\vspace{0.1cm}
\subsection{Verification Cost Proxy} \label{sec:eval-vcost}
The metrics above assess whether grounding is \emph{correct}; phrase-level attribution also aims to reduce how much source text a clinician must inspect. We therefore define a verification-cost proxy over \emph{correctly grounded phrases}. Let $\mathcal{P}^{\checkmark}$ be the predicted phrases that pass the provenance and faithfulness gates of \textbf{P-P}; for each instance with $\mathcal{P}^{\checkmark} \neq \emptyset$, let \textit{read} be the total number of source tokens in the admissible predicted phrases and $\operatorname{read}_{\text{sent}}$ the tokens in the cited sentences they come from:
\begin{equation}
\fontsize{8pt}{9pt}\selectfont
\textbf{Verification Cost Reduction (VCR)} = \frac{1}{|\mathcal{I}^{\checkmark}|}\!\sum_{i \in \mathcal{I}^{\checkmark}}\!\! \left(1 - \frac{\operatorname{read}(i)}{\operatorname{read}_{\text{sent}}(i)}\right),
\end{equation}
where $\mathcal{I}^{\checkmark}$ collects instances with at least one admissible phrase. \textbf{VCR} is $0$ when admissible phrases span their entire cited sentences and approaches $1$ as they narrow toward pinpoint evidence. Fabricated or unfaithful spans do not enter the count, as they are penalized by \textbf{P-P}, and missed evidence is handled by \textbf{C-R} and \textbf{P-R}. Because $\operatorname{read}$ grows with the highlighted span, \textbf{VCR} can be inflated by marking too few or too short phrases, so it is meaningful only alongside \textbf{P-R}, which captures evidence that is both narrow and complete. As a reference, reproducing the expert phrase set yields the annotation's own \textbf{VCR}; higher values indicate narrower spans, not a hard ceiling.

\section{Experiments and Results} \label{sec:experiment}
We organize the study around four questions. \textbf{RQ1}~(\S\ref{sec:exp-bench}): how well do current LLMs perform the task, and where does the difficulty lie? \textbf{RQ2}~(\S\ref{sec:exp-aspect}): how does difficulty vary across the sixteen aspects? \textbf{RQ3}~(\S\ref{sec:exp-scale}): how does grounding change with scale under a fixed protocol? \textbf{RQ4}~(\S\ref{sec:exp-metric}): do the proposed metrics align with human judgment, capture non-redundant dimensions, and remain robust to component and threshold choices?

\subsection{Experimental Setup} \label{sec:exp-setup}
\noindent\textbf{Models.} We benchmark eleven instruction-tuned LLMs across three tiers. \emph{Frontier proprietary}: GPT-5.2~\citep{openai2025gpt52}, Claude~Sonnet~4.6~\citep{anthropic2026sonnet46card}, Grok~4.3~\citep{xai2026grok43}, and Qwen3.7-Max~\citep{qwen2026qwen37max}. \emph{Large open-weight}: DeepSeek-V4-Pro~\citep{deepseek2026v4}, GLM-5.1~\citep{zai2026glm51}, and Qwen3.5-397B~\citep{qwen3.5}. \emph{Scale ladder}: Qwen3 dense models at $4$B, $8$B, $14$B, and $32$B~\citep{yang2025qwen3technicalreport}. The first two tiers test whether the capability probed by \textsc{MAPLE} is tied to closed weights or a particular model family, while the ladder isolates scale under a shared evaluation protocol.

\vspace{0.1cm}
\noindent\textbf{Configuration.} All models are queried 3-shot with a unified prompt that fixes the output schema: summary, cited sentence indices, contributory phrases, and abstention. A representative prompt is provided in the supplementary material. We use greedy decoding (temperature~$0$), so reported differences reflect cross-instance variability over the $1{,}799$ instances rather than decoding stochasticity. Open-weight models are either run locally on $6\times$NVIDIA~A6000 GPUs or accessed through the OpenRouter API, while proprietary models use official APIs with pinned snapshots. The model checkpoints are provided in the supplementary material. The evaluation backbone (\S\ref{sec:eval-prelim}) is shared across all models, so any backbone error is systematic rather than model-specific.

\vspace{0.1cm}
\noindent\textbf{Metrics.}
We report content metrics (\textbf{S-R}, \textbf{S-P}, \textbf{S-F1}), traceability metrics (\textbf{C-R}, \textbf{C-P}, \textbf{C-F1}, \textbf{C-A}), and locatability metrics (\textbf{P-R}, \textbf{P-P}, \textbf{P-F1}$_s$, \textbf{P-F1}$_h$), where \textbf{P-F1}$_s$ uses soft semantic matching and \textbf{P-F1}$_h$ uses hard ROUGE-1 matching \citep{lin-2004-rouge}. We also report verification-cost reduction (\textbf{VCR}); all metrics are averaged over the $1{,}799$ instances.

\newcolumntype{Y}{>{\centering\arraybackslash}X}

\begin{table*}[t]
\centering
\fontsize{9pt}{9.5pt}\selectfont
\setlength{\tabcolsep}{2.5pt}
\renewcommand{\arraystretch}{1.15}
\caption{\textbf{Main results on \textsc{MAPLE} (\%).}
We report metrics for content coverage, traceability, locatability, and VCR (soft semantic matching).
For locatability, P-F1$_s$ and P-F1$_h$ denote the soft semantic-matching and hard lexical-containment (ROUGE-1) variants, respectively. Best and second-best scores in each column are shown in \textbf{bold} and \underline{underlined}. Model checkpoints, a representative prompt, and results with 95\% confidence intervals are provided in the supplementary material.}
\vspace{-5pt}
\label{tab:main}

\begin{tabular*}{\textwidth}{@{\extracolsep{\fill}}l*{12}{c}@{}}
\toprule
& \multicolumn{3}{c}{\textbf{Content} (Summary)}
& \multicolumn{4}{c}{\textbf{Traceability} (Citation)}
& \multicolumn{4}{c}{\textbf{Locatability} (Phrase)}
& \\
\cmidrule(lr){2-4}\cmidrule(lr){5-8}\cmidrule(lr){9-12}
\textbf{Model}
& S-R & S-P & S-F1
& C-R & C-P & C-F1 & C-A
& P-R & P-P & P-F1$_s$ & P-F1$_h$
& \textbf{VCR} \\
\midrule

\multicolumn{13}{@{}l}{\textit{Frontier proprietary}} \\
\arrayrulecolor{gray}\hline\arrayrulecolor{black}
GPT-5.2           & 77.3 & 61.0 & 68.2 & 89.1 & 86.5 & 87.8 & 85.6 & 62.6 & 83.6 & 71.6 & 59.3 & 52.8 \\
Claude Sonnet 4.6 & \textbf{83.9} & 65.2 & 73.4 & \textbf{93.0} & 89.0 & \textbf{90.9} & 85.4 & 71.4 & 87.7 & 78.7 & 70.1 & 62.9 \\
Grok 4.3          & \underline{77.6} & 72.8 & 75.1 & 88.8 & \underline{90.6} & 89.7 & \underline{88.2} & 69.8 & \underline{89.5} & 78.4 & 71.8 & 67.5 \\
Qwen3.7-Max       & 75.5 & 71.8 & 73.6 & 91.2 & 87.6 & 89.4 & 84.5 & 73.6 & 88.4 & 80.4 & 73.6 & \textbf{70.5} \\

\arrayrulecolor{gray}\hline\arrayrulecolor{black}
\multicolumn{13}{@{}l}{\textit{Large open-weight}} \\
\arrayrulecolor{gray}\hline\arrayrulecolor{black}
DeepSeek-V4-Pro & 75.7 & \underline{77.1} & 76.4 & 89.7 & 87.6 & 88.6 & 86.0 & \underline{75.7} & 86.6 & \underline{80.8} & \underline{74.9} & 68.5 \\
GLM-5.1         & 76.1 & \textbf{77.9} & \textbf{77.0} & \underline{91.8} & 88.9 & 90.3 & 86.3 & \textbf{79.2} & \textbf{90.6} & \textbf{84.5} & \textbf{78.1} & \underline{70.2} \\
Qwen3.5-397B-A17B    & 75.5 & 71.7 & 73.5 & 90.0 & 87.5 & 88.7 & 84.7 & 72.5 & 86.7 & 79.0 & 71.3 & 67.8 \\

\arrayrulecolor{gray}\hline\arrayrulecolor{black}
\multicolumn{13}{@{}l}{\textit{Scale ladder (\textit{Qwen3} dense)}} \\
\arrayrulecolor{gray}\hline\arrayrulecolor{black}
Qwen3-32B & 77.4 & 76.2 & 76.8 & 90.9 & 89.4 & 90.1 & 86.7 & 71.5 & 82.7 & 76.7 & 69.1 & 67.7 \\
Qwen3-14B & 76.7 & \underline{77.1} & \underline{76.9} & 90.1 & \textbf{91.1} & \underline{90.6} & \textbf{88.8} & 72.9 & 85.3 & 78.6 & 72.4 & 68.1 \\
Qwen3-8B  & 70.3 & 76.8 & 73.4 & 91.7 & 82.6 & 86.9 & 83.0 & 63.0 & 69.5 & 66.1 & 58.8 & 67.8 \\
Qwen3-4B  & 69.5 & 74.9 & 72.1 & 85.9 & 86.5 & 86.2 & 83.8 & 61.1 & 73.1 & 66.6 & 59.2 & 66.1 \\
\bottomrule
\end{tabular*}
\vspace{-15pt}
\end{table*}

\subsection{RQ1: Benchmarking LLMs} \label{sec:exp-bench}
\noindent\textbf{Locatability remains the key bottleneck.}
Table~\ref{tab:main} shows that sentence-level traceability is consistently the easiest component: all models achieve high C-F1 ($86.2$--$90.9\%$). Content quality is lower and more variable (S-F1 $68.2$--$77.0\%$), while phrase-level grounding remains the most discriminative axis. Although the best systems reach strong P-F1 scores, locatability exhibits the widest spread across models ($66.1$--$84.5\%$, an $18.4$-point range, compared with $4.7$ points for C-F1). This suggests that current LLMs can often produce plausible summaries and identify supporting sentences, but their ability to isolate the exact phrases a clinician must inspect is less stable and remains the central challenge in \textsc{MAPLE}.

\vspace{0.1cm}
\noindent\textbf{The failure is conservative highlighting.}
The two grounding axes fail differently. Traceability recall and precision stay balanced (C-R $85.9$--$93.0\%$, C-P $82.6$--$91.1\%$), whereas locatability is consistently precision-heavy: every model posts a higher P-P than P-R, often widely (e.g., GPT-5.2 $83.6$ vs.\ $62.6\%$; Qwen3-4B $73.1$ vs.\ $61.1\%$). Models emit a small set of admissible spans but leave much of the reference set uncovered. Because \textbf{C-A} stays high ($83.0$--$88.8\%$), the failure is not orphan claims but evidence cited at the sentence level yet insufficiently localized within it.

\vspace{0.1cm}
\noindent\textbf{No model dominates.}
Leadership is split across axes: GLM-5.1 leads on content (S-F1 $77.0\%$) and locatability (P-F1 $84.5\%$), Claude~Sonnet~4.6 on traceability (C-F1 $90.9\%$), and Qwen3-14B on both C-P ($91.1\%$) and C-A ($88.8\%$). Qwen3.7-Max records the highest VCR ($70.5\%$), but at a lower P-R ($73.6\%$) than GLM-5.1's ($79.2\%$), so its evidence is partly narrower rather than more complete (\S\ref{sec:eval-vcost}). The proprietary tier does not dominate either: GPT-5.2 posts the lowest S-P ($61.0\%$) and VCR ($52.8\%$), along with the weakest P-R in its tier ($62.6\%$). With no uniform winner and the best phrase grounder (GLM-5.1) still missing roughly one in five reference phrases (P-R $79.2\%$), \textsc{MAPLE} remains unsaturated.

\vspace{0.1cm}
\noindent\textbf{Open weights are competitive on grounding.}
Large open-weight models match or exceed proprietary ones on phrase grounding: GLM-5.1 attains the best locatability (P-F1 $84.5\%$) and DeepSeek-V4-Pro ranks second (P-F1 $80.8\%$, P-R $75.7\%$), both surpassing every proprietary model on this axis. Proprietary models retain an edge only on a few recall-oriented measures, such as Claude's C-R ($93.0\%$) and S-R ($83.9\%$). The capability probed by \textsc{MAPLE} is thus not gated behind closed weights, which makes the benchmark actionable for open research.

\begin{figure}[t]
\centering
\begin{tikzpicture}
\definecolor{cS}{HTML}{FF637E}
\definecolor{cC}{HTML}{C71D36}
\definecolor{cP}{HTML}{A50C36}
\begin{axis}[
    width=\linewidth, height=4cm,
    ybar=0pt, bar width=4.5pt,
    ymin=50, ymax=108,
    ylabel={F1 (\%)},
    symbolic x coords={P,R,B,I,C,M,TD,OB,O,F,PE,SE,FD,AE,FU,RE},
    xtick=data, x tick label style={font=\footnotesize},
    legend style={at={(0.5,1.04)}, anchor=south, legend columns=3, font=\footnotesize, draw=none, column sep=10pt},
    ymajorgrids, grid style={gray!20},
    tick align=outside, xtick pos=bottom, ytick pos=left,
    enlarge x limits=0.04,
    clip=false,
]
\addplot[fill=cS, draw=none] coordinates {(P,66.7)(R,77.8)(B,82.2)(I,71.3)(C,56.4)(M,73.6)(TD,71.2)(OB,68.8)(O,62.9)(F,84.8)(PE,80.6)(SE,82.4)(FD,77.8)(AE,77.8)(FU,92.2)(RE,87.3)};
\addplot[fill=cC, draw=none] coordinates {(P,84.2)(R,95.2)(B,98.8)(I,92.1)(C,77.4)(M,87.3)(TD,97.8)(OB,73.2)(O,87.7)(F,94.9)(PE,89.2)(SE,89.9)(FD,96.3)(AE,89.6)(FU,74.6)(RE,97.4)};
\addplot[fill=cP, draw=none] coordinates {(P,72.3)(R,72.2)(B,91.2)(I,74.6)(C,76.7)(M,80.9)(TD,83.0)(OB,62.2)(O,66.8)(F,80.4)(PE,82.0)(SE,85.9)(FD,72.1)(AE,70.4)(FU,88.5)(RE,87.5)};
\legend{S-F1, C-F1, P-F1}
\node[font=\scriptsize, xshift=-4.5pt, yshift=4pt] at (axis cs:FU,92.2) {$\star$};
\node[font=\scriptsize, xshift=-4.5pt, yshift=4pt] at (axis cs:C,56.4)  {$\bullet$};
\node[font=\scriptsize, yshift=4pt] at (axis cs:B,98.8)  {$\star$};
\node[font=\scriptsize, yshift=4pt] at (axis cs:OB,73.2) {$\bullet$};
\node[font=\scriptsize, xshift=4.5pt, yshift=4pt] at (axis cs:B,91.2)  {$\star$};
\node[font=\scriptsize, xshift=4.5pt, yshift=4pt] at (axis cs:OB,62.2) {$\bullet$};
\node[font=\scriptsize\itshape, anchor=north] at ([yshift=-13pt]axis cs:R,50)  {Pop.\ \& Design};
\node[font=\scriptsize\itshape, anchor=north] at ([yshift=-13pt]axis cs:C,50)  {Intervention};
\node[font=\scriptsize\itshape, anchor=north] at ([yshift=-13pt]axis cs:F,50)  {Objective \& Outcomes};
\node[font=\scriptsize\itshape, anchor=north] at ([yshift=-13pt]axis cs:FU,50) {Safety \& Meta};
\end{axis}
\end{tikzpicture}
\vspace{-25pt}
\caption{Per-aspect F1 (\%) averaged over eleven models on the content, traceability, and locatability axes. Aspects are grouped by their role in RCT reporting (Table~\ref{tab:aspects}); $\star$ and $\bullet$ mark the best and worst aspect for each axis. Locatability is highest for short, near-verbatim evidence (B, FU, RE, SE) and lowest for diffuse or interpretive spans (OB, O, AE).}
\label{fig:per-aspect}
\vspace{-20pt}
\end{figure}

\subsection{RQ2: Per-Aspect Difficulty} \label{sec:exp-aspect}
We examine how difficulty varies across the sixteen aspects. Figure~\ref{fig:per-aspect} reports per-aspect F1 averaged over all eleven models, with $\star$ and $\bullet$ marking the best and worst aspect per axis, so the pattern reflects the task itself rather than any single model.

\vspace{0.1cm}
\noindent\textbf{Span concentration drives difficulty.}
Locatability varies by aspect but never collapses: even the hardest, \textit{Objective}, stays above $60\%$ P-F1 ($62.2\%$). Grounding is strongest when the evidence is short and near-verbatim, as in \textit{Blinding} (P-F1 $91.2\%$), \textit{Funding} ($88.5\%$), \textit{Registration} ($87.5\%$), and \textit{Secondary Endpoint} ($85.9\%$), and weakest when it is diffuse or interpretive, as in \textit{Objective} ($62.2\%$), \textit{Outcomes} ($66.8\%$), and \textit{Adverse Events} ($70.4\%$). The ordering tracks how concentrated the contributory span is rather than narrative complexity: \textit{Findings}, though a high-level synthesis aspect, still grounds well ($80.4\%$) because its evidence often sits in a single conclusory sentence.

\vspace{0.1cm}
\noindent\textbf{The bottleneck is aspect-dependent.}
On aspects with concentrated evidence, locatability often matches or exceeds content: \textit{Intervention} (P-F1 $74.6\%$ vs.\ S-F1 $71.3\%$), \textit{Comparator} ($76.7\%$ vs.\ $56.4\%$), and \textit{Participants} ($72.3\%$ vs.\ $66.7\%$), where the challenge lies in reconstructing the content rather than locating its phrases. The corpus-level finding thus holds on average, but the bottleneck shifts with aspect type: concentrated evidence narrows the gap, whereas interpretive aspects such as \textit{Objective} and \textit{Outcomes} degrade on both axes. Comparator also showed the lowest IAA ($\kappa=0.44$, \S\ref{sec:quality}), so its difficulty likely reflects both source ambiguity and reduced label reliability.

\subsection{RQ3: Scale Within a Fixed Evaluation Protocol} \label{sec:exp-scale}
\noindent\textbf{Scale improves locatability most.} The dense \textit{Qwen3} ladder isolates model size under a fixed evaluation protocol (Figure~\ref{fig:scale-ladder}, left). From $4$B to $32$B all three axes improve on net, but P-F1 gains the most: $+10.13$ points ($95\%$ BCa CI $[+8.64,+11.64]$), against $+4.71$ for S-F1 ($[+3.56,+5.86]$) and $+3.92$ for C-F1 ($[+3.01,+4.89]$). Though P-F1 dips slightly at $8$B before rising sharply at $14$B, a paired bootstrap confirms its net gain significantly exceeds both S-F1 ($+5.42$, $[+3.80,+7.07]$) and C-F1 ($+6.21$, $[+4.69,+7.71]$), while S-F1 and C-F1 improve at similar rates. Scaling therefore primarily relieves the locatability bottleneck of \S\ref{sec:exp-bench}.

\vspace{0.1cm}
\noindent\textbf{Gains saturate beyond $14$B.}
The trend is non-monotone (Figure~\ref{fig:scale-ladder}). From $14$B to $32$B, neither S-F1 ($-0.13$, $[-1.08,+0.88]$) nor C-F1 ($-0.43$, $[-1.16,+0.28]$) changes significantly, while P-F1 drops $1.92$ points ($[-3.11,-0.73]$), with falls in both P-P ($85.3\!\rightarrow\!82.7\%$) and P-R ($72.9\!\rightarrow\!71.5\%$). The largest model is thus not the best phrase grounder: $14$B is the sweet spot under this protocol, and the precision-heavy failure mode persists rather than disappearing at $32$B.

\begin{figure}[t]
\centering
\begin{tikzpicture}
\definecolor{cS}{HTML}{FF637E}
\definecolor{cC}{HTML}{C71D36}
\definecolor{cP}{HTML}{A50C36}
\begin{groupplot}[
    group style={group size=2 by 1, horizontal sep=2cm},
    width=0.5\linewidth, height=4cm,
    xmode=log, log basis x=2, xtick={4,8,14,32}, xticklabels={4,8,14,32},
    x tick label style={font=\footnotesize},
    xlabel={Parameters (B)},
    ymajorgrids, grid style={gray!20},
    tick align=outside, xtick pos=bottom, ytick pos=left,
    enlarge x limits=0.14,
]
\nextgroupplot[
    ylabel={F1 (\%)}, ymin=60, ymax=95,
    legend style={at={(0.5,1.1)}, anchor=south, legend columns=3, font=\scriptsize, draw=none, column sep=4pt},
]
\addplot[cS, mark=square*, thick, mark size=1.6pt] coordinates {(4,72.08)(8,73.43)(14,76.92)(32,76.79)};
\addplot[cC, mark=*, thick, mark size=1.6pt]        coordinates {(4,86.22)(8,86.92)(14,90.57)(32,90.14)};
\addplot[cP, mark=triangle*, thick, mark size=2pt]  coordinates {(4,66.56)(8,66.10)(14,78.61)(32,76.70)};
\legend{S-F1, C-F1, P-F1}
\nextgroupplot[
    ylabel={Locatability score (\%)}, ymin=55, ymax=90,
    legend style={at={(0.5,1.1)}, anchor=south, legend columns=2, font=\scriptsize, draw=none, column sep=8pt},
]
\addplot[cP, mark=*, thick, mark size=1.6pt] coordinates {(4,73.12)(8,69.52)(14,85.26)(32,82.68)};
\addplot[cP, mark=o, thick, mark size=1.8pt, densely dashed] coordinates {(4,61.09)(8,63.01)(14,72.92)(32,71.52)};
\legend{P-P, P-R}
\end{groupplot}
\end{tikzpicture}
\vspace{-25pt}
\caption{Scaling on the \textit{Qwen3} dense ladder ($4$B$\rightarrow$$32$B). \textbf{Left:} locatability gains most with scale ($+10.13$ points, more than twice the content or traceability gain), but the trend is non-monotone: P-F1 peaks at $14$B and regresses at $32$B. \textbf{Right:} both locatability precision (P-P) and recall (P-R) peak at $14$B, so the largest model is not the best phrase grounder.}
\label{fig:scale-ladder}
\vspace{-20pt}
\end{figure}

\subsection{RQ4: Meta-Evaluation of the Metrics} \label{sec:exp-metric}
We validate the framework along two dimensions: \emph{validity}, that is, whether the metrics measure intended properties without redundancy, and \emph{robustness}, that is, whether scores remain stable under component and threshold changes, examined through human correlation and threshold sweeps.

\vspace{0.3em}
\noindent\textit{\textbf{Validity.}}

\noindent\textbf{Human protocol.}
We sample $160$ GLM-5.1 instances ($10$ per aspect) and manually re-create the binary decisions underlying each learned-judge metric. C-R is excluded, as it reduces to exact sentence-set membership ($c \in \mathcal{C}'$). Two in-house annotators, distinct from both the dataset annotators (\S\ref{sec:protocol}) and quality reviewers (\S\ref{sec:quality}), independently judge content (S-R, S-P), traceability (C-P, C-A), and locatability (P-R, P-P) under the same reference visibility as the automatic definitions: recall-side tasks see the expert reference, reference-free tasks (C-P, C-A, P-P) do not. After calibration they reach $91.7\%$ agreement over $2{,}688$ judgments; disagreements are adjudicated to a single label and aggregated with the same formulas as the automatic metrics.

\vspace{0.1cm}
\noindent\textbf{The metrics track human judgment.}
Every automatic metric correlates strongly with its human counterpart (Figure~\ref{fig:meta-eval}a; Spearman $\rho$ over $160$ instances, all $p<10^{-18}$), with traceability closest (C-P/C-A $\rho=0.82/0.79$), content next (S-R/S-P $\rho=0.80/0.79$), and locatability lowest yet still strong (P-R/P-P $\rho=0.68/0.63$). Thus, although locatability is the hardest axis for both models and judges, its metric captures the intended human decision rather than a vague proxy.

\begin{figure}[t]
\centering
\begin{tikzpicture}[baseline=(current axis.north)]
\definecolor{cl}{HTML}{FF2056}
\begin{axis}[
    width=0.5\linewidth, height=3.8cm,
    ybar, bar width=8pt, ymin=0.5, ymax=1,
    ylabel={$\rho$},
    symbolic x coords={SR,SP,CP,CA,PR,PP},
    xtick=data,
    xticklabels={S-R,S-P,C-P,C-A,P-R,P-P},
    x tick label style={font=\tiny, rotate=35, anchor=east},
    enlarge x limits=0.09,
    nodes near coords,
    nodes near coords style={font=\tiny, rotate=90, anchor=west, /pgf/number format/fixed, /pgf/number format/precision=2},
    title style={font=\footnotesize}, title={(a)},
    axis on top, tick align=outside, xtick pos=bottom, ytick pos=left,
]
\addplot[fill=cl, draw=none] coordinates {(SR,0.80)(SP,0.79)(CP,0.82)(CA,0.79)(PR,0.68)(PP,0.63)};
\draw[gray!40] (rel axis cs:0,0) rectangle (rel axis cs:1,1);
\end{axis}
\end{tikzpicture}
\hfill
\begin{tikzpicture}[baseline=(current axis.north)]
\definecolor{cr}{HTML}{155DFC}
\begin{axis}[
    width=0.5\linewidth, height=3.8cm,
    ybar, bar width=3.8pt, ymin=0, ymax=18,
    ylabel={P-F1 lift (pts)},
    symbolic x coords={B,SE,FU,F,M,RE,AE,TD,PE,I,P,C,O,FD,OB,R},
    xtick=data, x tick label style={font=\tiny},
    enlarge x limits=0.04,
    title style={font=\footnotesize}, title={(b)},
    axis on top, tick align=outside, xtick pos=bottom, ytick pos=left,
]
\addplot[fill=cr, draw=none] coordinates {(B,0.8)(SE,2.7)(FU,4.0)(F,4.3)(M,5.6)(RE,6.0)(AE,6.1)(TD,7.0)(PE,7.0)(I,7.2)(P,7.6)(C,7.7)(O,8.9)(FD,12.2)(OB,13.9)(R,15.1)};
\draw[gray!40] (rel axis cs:0,0) rectangle (rel axis cs:1,1);
\end{axis}
\end{tikzpicture}
\vspace{-10pt}
\caption{Validity of the metrics. \textbf{(a)} Spearman $\rho$ between automatic metrics and human judgments over $160$ GLM-5.1 instances. \textbf{(b)} Per-aspect P-F1 lift from lexical ROUGE-1 to semantic cosine matching, pooled across eleven models. Gains concentrate on interpretive aspects (R, OB, FD, O) and vanish on near-verbatim ones (B, SE, FU).}
\label{fig:meta-eval}
\vspace{-20pt}
\end{figure}

\vspace{0.1cm}
\noindent\textbf{Semantic matching outperforms lexical overlap.}
To isolate semantic phrase matching, we replace the encoder with ROUGE-1 F1 and re-evaluate locatability under the same provenance, faithfulness, and abstention settings. Across all $1{,}799 \times 11$ instances, lexical matching lowers P-F1 by $7.5$ points on average, with over $90\%$ of the loss coming from P-R. The gap is largest for aspects that require interpretation or paraphrase, namely \textit{Randomization} ($+15.1$ pts), \textit{Objective} ($+13.9$), and \textit{Follow-Up Duration} ($+12.2$), and smallest for near-verbatim spans, namely \textit{Blinding} ($+0.8$), \textit{Secondary Endpoint} ($+2.7$), and \textit{Funding} ($+4.0$) (Figure~\ref{fig:meta-eval}b). Semantic similarity thus recovers paraphrases that lexical overlap misses while preserving the model ranking (Spearman $\rho=0.95$).

\vspace{0.1cm}
\noindent\textbf{The axes measure distinct dimensions.}
Pairwise Pearson correlations among the seven base metrics stay low (Figure~\ref{fig:rq-robustness}a, pooled over $19{,}789$ instances). The sole exception is C-P with C-A ($0.84$), an expected pair that reflects complementary views of traceability. Excluding it, every off-diagonal pair is at most $0.40$ (the maximum being S-R with C-R), and $76\%$ are at most $0.33$; even P-R and P-P correlate only at $0.33$, confirming that the metrics capture related but distinct failure modes.

\begin{table}[t]
\centering\small
\setlength{\tabcolsep}{6.5pt}
\renewcommand{\arraystretch}{1.15}
\caption{Robustness to swappable backbone components. Each row replaces one default component and reports Spearman rank correlation and mean absolute score change on the affected metrics. Model rankings remain stable, with a maximum position shift of $2$; the decomposer swap also tests same-family affinity using Gemma-4-31B.}

\vspace{-5pt}
\label{tab:robust}
\begin{tabular}{llccc}
\toprule
\textbf{Component} & \textbf{Alternative} & \textbf{Metrics} & \textbf{Rank $\rho$} & \textbf{Mean $|\Delta|$ (pts)} \\
\midrule
Decomposer $\mathcal{E}$ & \texttt{gemma-4-31b} & S-F1 / C-F1 / C-A & $0.96$ / $0.96$ / $0.96$ & $1.15$ / $2.24$ / $1.47$ \\
Entailment $\phi$ & \texttt{deberta-v3-large} & S-F1 / C-F1 / C-A & $0.87$ / $0.99$ / $0.96$ & $1.74$ / $0.55$ / $2.34$ \\
Encoder $\mathcal{S}$ & \texttt{bge-large-en-v1.5} & P-F1 & $0.91$ & $9.23$ \\
\bottomrule
\end{tabular}
\vspace{-15pt}
\end{table}

\vspace{0.3em}
\noindent\textit{\textbf{Robustness.}}

\noindent\textbf{Rankings survive backbone substitutions.}
We replace each swappable component (decomposer $\mathcal{E}$, entailment judge $\phi$, encoder $\mathcal{S}$) with an alternative from a different model family and re-score all eleven models (Table~\ref{tab:robust}); the rankings stay stable. Replacing TRUE with a DeBERTa-v3 NLI cross-encoder preserves traceability tightly ($\rho=0.99$ on C-F1, $0.96$ on C-A) and content reasonably well ($0.87$ on S-F1). Replacing MiniLM with BGE-large-en preserves the P-F1 ranking broadly ($\rho=0.91$), though absolute scores rise by $9.2$ points, reflecting calibration rather than relative quality. Replacing the Qwen3.6-35B-A3B decomposer with Gemma-4-31B leaves every decomposer-driven ranking tight ($\rho\ge0.95$), with the \textit{Qwen3} ladder unchanged. The reported comparisons thus reflect grounding ability rather than affinity to any particular backbone.

\begin{figure}[t]
\centering
\begin{tikzpicture}[baseline=(current axis.north)]
\begin{axis}[
    width=0.36\linewidth, height=3.8cm,
    enlargelimits=false,
    colormap={whitered}{color=(white) color=(red!70!black)},
    xlabel={(a)},
    point meta min=0, point meta max=1,
    xtick={0,1,2,3,4,5,6}, ytick={0,1,2,3,4,5,6},
    xticklabels={S-R,S-P,C-R,C-P,C-A,P-R,P-P},
    yticklabels={S-R,S-P,C-R,C-P,C-A,P-R,P-P},
    x tick label style={font=\tiny, rotate=45, anchor=east},
    y tick label style={font=\tiny},
    y dir=reverse,
    nodes near coords={\pgfmathprintnumber[fixed,precision=2]\pgfplotspointmeta},
    nodes near coords style={font=\tiny, anchor=center},
]
\addplot[matrix plot*, mesh/cols=7, point meta=explicit] table[meta=val] {
x y val
0 0 1.00
1 0 0.22
2 0 0.40
3 0 0.23
4 0 0.25
5 0 0.29
6 0 0.36
0 1 0.22
1 1 1.00
2 1 0.27
3 1 0.30
4 1 0.33
5 1 0.13
6 1 0.23
0 2 0.40
1 2 0.27
2 2 1.00
3 2 0.31
4 2 0.36
5 2 0.39
6 2 0.28
0 3 0.23
1 3 0.30
2 3 0.31
3 3 1.00
4 3 0.84
5 3 0.15
6 3 0.32
0 4 0.25
1 4 0.33
2 4 0.36
3 4 0.84
4 4 1.00
5 4 0.18
6 4 0.32
0 5 0.29
1 5 0.13
2 5 0.39
3 5 0.15
4 5 0.18
5 5 1.00
6 5 0.33
0 6 0.36
1 6 0.23
2 6 0.28
3 6 0.32
4 6 0.32
5 6 0.33
6 6 1.00
};
\end{axis}
\end{tikzpicture}
\hfill
\begin{tikzpicture}[baseline=(current axis.north)]
\begin{axis}[
    width=0.36\linewidth, height=3.8cm,
    xlabel={(b) $\tau$ ($\eta{=}0.80$)}, ylabel={mean P-F1 (\%)},
    ymin=60, ymax=90,
    xmin=0.58, xmax=0.82,
    xtick={0.60,0.65,0.70,0.75,0.80},
    x tick label style={font=\tiny, rotate=30, anchor=east},
    y tick label style={font=\tiny},
    label style={font=\scriptsize},
    ymajorgrids, grid style={gray!20},
    tick align=outside, xtick pos=bottom, ytick pos=left,
]
\addplot+[mark=*, blue!70, thick, mark size=1.8pt]
    coordinates {(0.60,80.89)(0.65,78.85)(0.70,76.49)(0.75,73.63)(0.80,71.43)};
\draw[gray!40] (rel axis cs:0,0) rectangle (rel axis cs:1,1);
\end{axis}
\end{tikzpicture}
\hfill
\begin{tikzpicture}[baseline=(current axis.north)]
\begin{axis}[
    width=0.36\linewidth, height=3.8cm,
    xlabel={(c) $\eta$ ($\tau{=}0.70$)}, ylabel={mean P-F1 (\%)},
    ymin=60, ymax=90,
    xmin=0.55, xmax=1.0,
    xtick={0.60,0.70,0.80,0.90,1.0},
    x tick label style={font=\tiny, rotate=30, anchor=east},
    y tick label style={font=\tiny},
    label style={font=\scriptsize},
    ymajorgrids, grid style={gray!20},
    tick align=outside, xtick pos=bottom, ytick pos=left,
]
\addplot+[mark=square*, teal!60!black, thick, mark size=1.6pt]
    coordinates {(0.60,78.19)(0.70,77.22)(0.80,76.49)(0.90,75.17)(0.95,74.96)};
\draw[gray!40] (rel axis cs:0,0) rectangle (rel axis cs:1,1);
\end{axis}
\end{tikzpicture}
\vspace{-10pt}
\caption{Robustness of the evaluation framework. \textbf{(a)} Pairwise Pearson correlations among the seven base metrics (pooled over $19{,}789$ instances); all off-diagonal pairs are at most $0.40$ except C-P / C-A ($0.84$). \textbf{(b,c)} Mean P-F1 when sweeping $\tau$ ($\eta{=}0.80$) and $\eta$ ($\tau{=}0.70$); thresholds shift absolute scores but preserve rankings throughout (Spearman $\rho\ge0.97$).}
\label{fig:rq-robustness}
\vspace{-25pt}
\end{figure}

\vspace{0.1cm}
\noindent\textbf{The thresholds are not load-bearing.}
The locatability axis uses two scalar thresholds, $\tau$ for semantic matching and $\eta$ for the fraction of phrase tokens matched by summary windows. Sweeping $\tau\in\{0.60,0.65,0.70,0.75,0.80\}$ at $\eta{=}0.80$ and $\eta\in\{0.60,0.70,0.80,0.90,0.95\}$ at $\tau{=}0.70$ (Figure~\ref{fig:rq-robustness}b,c), we find the rankings stable throughout: Spearman $\rho\ge0.97$ everywhere, no model moves more than two positions, and GLM-5.1 stays first with Qwen3-\{4B,8B\} the bottom pair. Absolute P-F1 shifts with $\tau$ (a $9.5$-point swing) more than with $\eta$ ($3.2$ points). Relative comparisons are thus robust to threshold choice, even if absolute scores depend on calibration, so practitioners need not tune these thresholds per dataset.

\section{Conclusion}
We introduced \textsc{MAPLE}, a benchmark for medical aspect-based summarization with phrase-level context attribution. It contains 152 RCT abstracts, 16 clinical aspects, and 1,799 human-annotated summaries grounded in cited sentences and contributory phrases, with double annotation and independent review on stratified subsets. Together with a decoupled evaluation framework, our eleven-LLM study shows that sentence-level traceability is relatively strong, while phrase-level locatability remains the more discriminative bottleneck. Scaling improves locatability more than content or traceability under our protocol, but the gains eventually saturate and may even regress, suggesting that precise evidence localization is not solved by scale alone. \textsc{MAPLE} aims to support clinical summarization systems whose outputs can be audited quickly, not merely read. The MAPLE annotated data, annotation guidelines, evaluation code, and derived metadata are released for research use. The annotations and metadata are released under CC BY 4.0, and the evaluation code under Apache 2.0. We will actively maintain the resource with expanded coverage in future releases.

\section*{Limitations}
\noindent\textbf{Scope of evaluation.}
We evaluate general-purpose LLMs rather than medical domain-adapted checkpoints such as OpenBioLLM, BioMistral, or Meditron; a controlled comparison would require domain-adapted models built on base models comparable to ours, which are not yet consistently available in the open-weight ecosystem. We leave such a comparison to future work, once matched open-weight base models become available.

\vspace{0.1cm}
\noindent\textbf{Corpus scope.}
\textsc{MAPLE} focuses on melanoma RCT abstracts to keep clinical guidance consistent and document structure comparable (\S\ref{sec:source}). Abstracts are the primary entry point for literature search and trial screening, so phrase-level grounding at this level is already practically useful; we leave generalization to other therapeutic areas, study designs, and full-text articles, where evidence may span many sections, to future work.

\vspace{0.1cm}
\noindent\textbf{Annotation coverage.}
The $20$-article double-annotated subset shows strong agreement ($\kappa=0.80$ on sentence selection), and human review of $50$ abstracts confirms label quality (\S\ref{sec:quality}), but the remaining $132$ abstracts were single-annotated. We treat this convergent evidence as supporting consistency at scale, while acknowledging single-annotator coverage of most of the corpus.

\vspace{0.1cm}
\noindent\textbf{Evaluation backbone.}
The framework relies on learned components (decomposer, entailment judge, encoder), so absolute scores reflect their calibration. Our meta-evaluation (\S\ref{sec:exp-metric}) shows that rankings stay stable under backbone and threshold changes, but cross-paper comparisons should emphasize relative rather than absolute scores.

\bibliography{acml26, mybibliography, anthology-1, anthology-2}

\clearpage

\appendix

\newcommand{\cistack}[3]{\shortstack[c]{#1\\[-1pt]{\tiny\color{gray}{[#2,#3]}}}}
\newcommand{\bestcistack}[3]{\shortstack[c]{\textbf{#1}\\[-1pt]{\tiny\color{gray}{[#2,#3]}}}}
\newcommand{\secondcistack}[3]{\shortstack[c]{\underline{#1}\\[-1pt]{\tiny\color{gray}{[#2,#3]}}}}

\begin{sidewaystable*}[p]

{\large\bfseries Supplementary Material 1: Full Results with 95\% Confidence Intervals\par}
\vspace{6pt}

\fontsize{8pt}{12pt}\selectfont
\setlength{\tabcolsep}{3.0pt}
\renewcommand{\arraystretch}{1.15}
\resizebox{0.97\textheight}{!}{%
\begin{tabular}{l ccc cccc ccc c}
\toprule
& \multicolumn{3}{c}{\textbf{Content} (Summary)}
& \multicolumn{4}{c}{\textbf{Traceability} (Citation)}
& \multicolumn{3}{c}{\textbf{Locatability} (Phrase)}
& \\
\cmidrule(lr){2-4}\cmidrule(lr){5-8}\cmidrule(lr){9-11}
\textbf{Model}
& S-R & S-P & S-F1
& C-R & C-P & C-F1 & C-A
& P-R & P-P & P-F1$_s$
& \textbf{VCR} \\
\midrule

\multicolumn{12}{@{}l}{\textit{Frontier proprietary}} \\
\arrayrulecolor{gray}\hline\arrayrulecolor{black}
GPT-5.2
& \cistack{77.3}{75.8}{79.0}
& \cistack{61.0}{59.2}{62.8}
& \cistack{68.2}{66.7}{69.6}
& \cistack{89.1}{87.7}{90.3}
& \cistack{86.5}{85.1}{87.9}
& \cistack{87.8}{86.5}{88.9}
& \cistack{85.6}{84.2}{86.9}
& \cistack{62.6}{60.8}{64.5}
& \cistack{83.6}{82.2}{84.9}
& \cistack{71.6}{70.2}{73.0}
& \cistack{52.8}{51.3}{54.1} \\

Claude Sonnet 4.6
& \bestcistack{83.9}{82.5}{85.1}
& \cistack{65.2}{63.6}{66.9}
& \cistack{73.4}{72.1}{74.6}
& \bestcistack{93.0}{92.0}{93.9}
& \cistack{89.0}{87.8}{90.1}
& \bestcistack{90.9}{90.1}{91.7}
& \cistack{85.4}{84.1}{86.6}
& \cistack{71.4}{69.7}{73.0}
& \cistack{87.7}{86.7}{88.7}
& \cistack{78.7}{77.5}{79.8}
& \cistack{62.9}{62.0}{64.0} \\

Grok 4.3
& \secondcistack{77.6}{76.0}{79.2}
& \cistack{72.8}{71.1}{74.5}
& \cistack{75.1}{73.8}{76.4}
& \cistack{88.8}{87.6}{90.0}
& \secondcistack{90.6}{89.3}{91.7}
& \cistack{89.7}{88.7}{90.7}
& \secondcistack{88.2}{87.0}{89.4}
& \cistack{69.8}{68.0}{71.5}
& \secondcistack{89.5}{88.3}{90.6}
& \cistack{78.4}{77.1}{79.7}
& \cistack{67.5}{66.4}{68.5} \\

Qwen3.7-Max
& \cistack{75.5}{73.9}{77.0}
& \cistack{71.8}{70.1}{73.5}
& \cistack{73.6}{72.3}{74.9}
& \cistack{91.2}{90.1}{92.3}
& \cistack{87.6}{86.3}{88.9}
& \cistack{89.4}{88.4}{90.3}
& \cistack{84.5}{83.2}{85.8}
& \cistack{73.6}{72.0}{75.3}
& \cistack{88.4}{87.3}{89.5}
& \cistack{80.4}{79.2}{81.5}
& \bestcistack{70.5}{69.5}{71.4} \\

\arrayrulecolor{gray}\hline\arrayrulecolor{black}
\multicolumn{12}{@{}l}{\textit{Large open-weight}} \\
\arrayrulecolor{gray}\hline\arrayrulecolor{black}
DeepSeek-V4-Pro
& \cistack{75.7}{74.1}{77.3}
& \secondcistack{77.1}{75.5}{78.6}
& \cistack{76.4}{75.0}{77.6}
& \cistack{89.7}{88.5}{90.8}
& \cistack{87.6}{86.2}{88.9}
& \cistack{88.6}{87.5}{89.7}
& \cistack{86.0}{84.6}{87.3}
& \secondcistack{75.7}{74.2}{77.2}
& \cistack{86.6}{85.3}{87.7}
& \secondcistack{80.8}{79.6}{81.9}
& \cistack{68.5}{67.5}{69.5} \\

GLM-5.1
& \cistack{76.1}{74.5}{77.6}
& \bestcistack{77.9}{76.4}{79.5}
& \bestcistack{77.0}{75.8}{78.2}
& \secondcistack{91.8}{90.7}{92.8}
& \cistack{88.9}{87.6}{90.1}
& \cistack{90.3}{89.3}{91.2}
& \cistack{86.3}{84.9}{87.5}
& \bestcistack{79.2}{77.6}{80.6}
& \bestcistack{90.6}{89.6}{91.5}
& \bestcistack{84.5}{83.4}{85.5}
& \secondcistack{70.2}{69.2}{71.1} \\

Qwen3.5-397B-A17B
& \cistack{75.5}{73.9}{77.0}
& \cistack{71.7}{70.0}{73.3}
& \cistack{73.5}{72.2}{74.8}
& \cistack{90.0}{88.8}{91.1}
& \cistack{87.5}{86.1}{88.8}
& \cistack{88.7}{87.6}{89.7}
& \cistack{84.7}{83.3}{86.0}
& \cistack{72.5}{70.8}{74.2}
& \cistack{86.7}{85.5}{87.9}
& \cistack{79.0}{77.7}{80.2}
& \cistack{67.8}{66.8}{68.8} \\

\arrayrulecolor{gray}\hline\arrayrulecolor{black}
\multicolumn{12}{@{}l}{\textit{Scale ladder (\textit{Qwen3} dense)}} \\
\arrayrulecolor{gray}\hline\arrayrulecolor{black}
Qwen3-32B
& \cistack{77.4}{75.8}{78.8}
& \cistack{76.2}{74.6}{77.7}
& \cistack{76.8}{75.5}{78.0}
& \cistack{90.9}{89.7}{91.9}
& \cistack{89.4}{88.1}{90.6}
& \cistack{90.1}{89.2}{91.0}
& \cistack{86.7}{85.4}{87.9}
& \cistack{71.5}{69.9}{73.2}
& \cistack{82.7}{81.3}{83.9}
& \cistack{76.7}{75.4}{78.0}
& \cistack{67.7}{66.8}{68.7} \\

Qwen3-14B
& \cistack{76.7}{75.1}{78.2}
& \secondcistack{77.1}{75.5}{78.6}
& \secondcistack{76.9}{75.6}{78.1}
& \cistack{90.1}{88.8}{91.1}
& \bestcistack{91.1}{89.9}{92.2}
& \secondcistack{90.6}{89.6}{91.4}
& \bestcistack{88.8}{87.5}{89.9}
& \cistack{72.9}{71.3}{74.5}
& \cistack{85.3}{83.9}{86.5}
& \cistack{78.6}{77.4}{79.8}
& \cistack{68.1}{67.1}{69.1} \\

Qwen3-8B
& \cistack{70.3}{68.7}{72.0}
& \cistack{76.8}{75.3}{78.3}
& \cistack{73.4}{72.2}{74.7}
& \cistack{91.7}{90.7}{92.7}
& \cistack{82.6}{81.2}{84.0}
& \cistack{86.9}{86.0}{87.9}
& \cistack{83.0}{81.6}{84.4}
& \cistack{63.0}{61.2}{64.8}
& \cistack{69.5}{67.8}{71.2}
& \cistack{66.1}{64.6}{67.6}
& \cistack{67.8}{66.6}{69.0} \\

Qwen3-4B
& \cistack{69.5}{67.7}{71.2}
& \cistack{74.9}{73.2}{76.6}
& \cistack{72.1}{70.7}{73.5}
& \cistack{85.9}{84.5}{87.2}
& \cistack{86.5}{85.2}{87.8}
& \cistack{86.2}{85.1}{87.3}
& \cistack{83.8}{82.5}{85.2}
& \cistack{61.1}{59.2}{63.0}
& \cistack{73.1}{71.4}{74.8}
& \cistack{66.6}{65.0}{68.1}
& \cistack{66.1}{64.8}{67.4} \\
\bottomrule
\end{tabular}%
}

\vspace{4pt}
\caption{\textsc{MAPLE} results with 95\% BCa bootstrap confidence intervals. Each cell shows the mean score on the first line and the confidence interval below, all in \%. Best and second-best scores in each column are shown in \textbf{bold} and \underline{underlined}.}
\label{tab:ci}
\end{sidewaystable*}

\clearpage

\section*{Supplementary Material 2: Aspect-Oriented Extraction Prompt}
\label{appendix:prompt}
\begin{promptbox}
You are an expert biomedical information extraction assistant.

You are given a medical abstract represented as a list of indexed sentences. Your task is to identify the sentence or sentences that explicitly describe the intervention used in the study, extract the key intervention-related phrases exactly as they appear in the source sentences, and generate a concise summary.

Target aspect:
Intervention

Definition:
The intervention refers to the experimental treatment, drug, regimen, procedure, or therapeutic strategy administered to the study participants. Relevant information may include the treatment name, drug dose, route of administration, dosing schedule, combination regimen, and treatment duration when these details are part of the experimental arm.

Instructions:

1. Sentence Selection
   Select only the sentence indices that explicitly describe the intervention.
   Prioritize sentences that provide concrete treatment details, such as drug names, dosage, route, schedule, or regimen composition.
   Do not select sentences that only describe the study background, objective, participant eligibility, outcomes, conclusions, or control treatment unless they directly specify the experimental intervention.
   If no sentence explicitly describes the intervention, return an empty list for "sentences", an empty string for "summary", and an empty list for "kps".

2. Key Phrase Extraction
   From the selected sentence or sentences, extract short and meaningful key phrases that directly support the intervention description.
   Each key phrase must be copied verbatim from the corresponding source sentence.
   Each key phrase must be an exact, contiguous substring of the sentence from which it is extracted.
   Do not paraphrase, normalize, expand abbreviations, or merge words from different sentences.
   Do not copy an entire sentence as one key phrase.
   Every sentence index appearing in "kps" must also appear in the "sentences" list.

3. Summary Generation
   Write one concise sentence summarizing the intervention.
   The summary may paraphrase the selected evidence, but it must remain faithful to the original abstract.
   If the intervention is absent, the summary must be an empty string.

4. Output Requirement
   Return only a valid JSON object.
   Do not include explanations, comments, markdown formatting, or any text outside the JSON object.

Output format:

{
  "sentences": [index1, index2],
  "summary": "One concise sentence summarizing the intervention.",
  "kps": [
    {
      "sentence": index1,
      "kps": [
        "verbatim intervention-related phrase",
        "another verbatim phrase"
      ]
    },
    {
      "sentence": index2,
      "kps": [
        "verbatim phrase from sentence index2"
      ]
    }
  ]
}

If the target aspect is absent, use the following format:

{
  "sentences": [],
  "summary": "",
  "kps": []
}

Example input:

Abstract:
[
  "0: Background: Advanced non-small-cell lung cancer remains difficult to treat after progression on platinum-based chemotherapy.",
  "1: We conducted a randomized phase II trial to evaluate the efficacy and safety of pembrolizumab combined with lenvatinib in patients with previously treated advanced non-small-cell lung cancer.",
  "2: Eligible patients were randomly assigned to receive pembrolizumab 200 mg intravenously every 3 weeks plus lenvatinib 20 mg orally once daily.",
  "3: The control group received docetaxel 75 mg/m2 intravenously every 3 weeks.",
  "4: The primary endpoint was progression-free survival, and secondary endpoints included overall survival, objective response rate, and treatment-related adverse events.",
  "5: Median progression-free survival was longer in the pembrolizumab-lenvatinib group than in the docetaxel group.",
  "6: Grade 3 or higher treatment-related adverse events occurred in 28\% of patients receiving pembrolizumab plus lenvatinib."
]

Expected output:

{
  "sentences": [2],
  "summary": "The experimental intervention was pembrolizumab 200 mg intravenously every 3 weeks combined with lenvatinib 20 mg orally once daily.",
  "kps": [
    {
      "sentence": 2,
      "kps": [
        "pembrolizumab 200 mg intravenously every 3 weeks",
        "lenvatinib 20 mg orally once daily"
      ]
    }
  ]
}

The remaining two examples are omitted...

Now perform the task on the following abstract.

Abstract:
{indexed_abstract}
\end{promptbox}

\clearpage
\section*{Supplementary Material 3: Evaluated Model Checkpoints}

\label{app:checkpoints}
\begin{table}[h]
\centering
\small
\begin{threeparttable}
\caption{Model checkpoints evaluated in \textsc{MAPLE}. Proprietary models are
queried through their official APIs with pinned snapshots; the large open-weight
models are accessed through the OpenRouter API, and the Qwen3 dense ladder is run
locally. Activated parameters per token are given in parentheses for
mixture-of-experts (MoE) models.}
\label{tab:checkpoints}
\begin{tabular}{lllll}
\toprule
Model & Provider & Access & Params\tnote{a} & Checkpoint / Model ID \\
\midrule
\addlinespace[3pt]
\multicolumn{5}{l}{\textit{Frontier proprietary}} \\
\addlinespace[3pt]
GPT-5.2           & OpenAI    & API        & ---             & \texttt{gpt-5.2-2025-12-11} \\
Claude Sonnet 4.6 & Anthropic & API        & ---             & \texttt{claude-sonnet-4-6}\tnote{b} \\
Grok 4.3          & xAI       & API        & ---             & \texttt{grok-4.3} \\
Qwen3.7-Max       & Alibaba   & API        & ---             & \texttt{qwen3.7-max-2026-05-20} \\
\hline
\addlinespace[3pt]
\multicolumn{5}{l}{\textit{Large open-weight (via OpenRouter)\tnote{c}}} \\
\addlinespace[3pt]

DeepSeek-V4-Pro & DeepSeek & OpenRouter & 1.6T (49B act.) & \texttt{deepseek/deepseek-v4-pro} \\
GLM-5.1         & Z.ai     & OpenRouter & 754B (MoE)      & \texttt{z-ai/glm-5.1} \\
Qwen3.5-397B    & Alibaba  & OpenRouter & 397B (17B act.) & \texttt{qwen/qwen3.5-397b-a17b} \\
\hline
\addlinespace[3pt]
\multicolumn{5}{l}{\textit{Scale ladder (Qwen3 dense, run locally)\tnote{d}}} \\
\addlinespace[3pt]

Qwen3-32B & Alibaba & Local & 32B & \texttt{Qwen/Qwen3-32B} \\
Qwen3-14B & Alibaba & Local & 14B & \texttt{Qwen/Qwen3-14B} \\
Qwen3-8B  & Alibaba & Local & 8B  & \texttt{Qwen/Qwen3-8B} \\
Qwen3-4B  & Alibaba & Local & 4B  & \texttt{Qwen/Qwen3-4B} \\
\bottomrule
\end{tabular}
\begin{tablenotes}[flushleft]
\footnotesize
\item[a] Total parameters, with activated parameters per token in parentheses for
MoE models; dashes denote undisclosed proprietary model sizes.
\item[b] The model ID \texttt{claude-sonnet-4-6} refers to a fixed
snapshot rather than an automatically updated alias.
\item[c] Queried through the OpenRouter API; the underlying open weights are
\texttt{deepseek-ai/DeepSeek-V4-Pro}, \texttt{zai-org/GLM-5.1}, and
\texttt{Qwen/Qwen3.5-397B-A17B} on Hugging Face.
\item[d] Run locally on 6$\times$NVIDIA A6000 GPUs.
\end{tablenotes}
\end{threeparttable}
\end{table}

\end{document}